# Open-Vocabulary Semantic Segmentation with Uncertainty Alignment for Robotic Scene Understanding in Indoor Building Environments

Yifan Xu[1], Vineet Kamat[2], and Carol Menassa[3]

[1]Ph.D. Candidate, Department of Civil and Environmental Engineering, University of Michigan, Ann Arbor, MI, USA, Email: yfx@umich.edu

[2]Professor, Department of Civil and Environmental Engineering, University of Michigan, Ann Arbor, MI, USA, Email: vkamat@umich.edu

[3]Professor, Department of Civil and Environmental Engineering, University of Michigan, Ann Arbor, MI, USA, Email: menassa@umich.edu

## ABSTRACT

The global rise in the number of people with physical disabilities, in part due to improvements in post-trauma survivorship and longevity, has amplified the demand for advanced assistive technologies to improve mobility and independence. Autonomous assistive robots, such as smart wheelchairs, require robust capabilities in spatial segmentation and semantic recognition to navigate complex built environments effectively. Place segmentation involves delineating spatial regions like rooms or functional areas, while semantic recognition assigns semantic labels to these regions, enabling accurate localization to user-specific needs. Existing approaches often utilize deep learning; however, these close-vocabulary detection systems struggle to interpret intuitive and casual human instructions. Additionally, most existing methods ignore the uncertainty of the scene recognition problem, leading to low success rates, particularly in ambiguous and complex environments. To address these challenges, we propose an open-vocabulary scene semantic segmentation and detection pipeline leveraging Vision Language Models (VLMs) and Large Language Models (LLMs). Our approach follows a "Segment-Detect-Select" framework for open-vocabulary scene classification, enabling adaptive and intuitive navigation for assistive robots in built environments. The technical approach consists of three different modules: 1) segment different rooms/regions of the scene; 2) leverage VLM to get the similarity score between people's descriptions and rooms; and 3) use adaptive conformal prediction (ACP) to select rooms according to the similarity scores to balance the uncertainty of the system. The evaluation results show that our framework can outperform the current state-of-the-art (SoTA) open-vocabulary scene segmentation, classification, and selection algorithms on the widely used Matterport3D (MP3D) real, building-scale scene datasets.

## INTRODUCTION

The global prevalence of disability and longevity has increased, leading to a growing demand for advanced assistive technologies to improve mobility and independence of individuals in indoor built environments (Alqahtani et al. 2019; Maiya et al. 2019; Rwegoshora et al. 2022). Among these, autonomous assistive robots, such as smart wheelchairs, have emerged as promising solutions to address the challenges faced in living and working within complex built environments (Misaros et al. 2023). Achieving effective robot navigation and autonomy necessitates robust spatial segmentation and semantic recognition capabilities, which are critical for scene understanding and interacting with dynamic artifacts, particularly those encountered indoors (Alqobali et al. 2023).

Spatial segmentation involves delineating distinct regions within an environment, such as identifying different rooms, corridors, or functional zones (Wang et al. 2019a), which has attracted increasing research attention in the robotics and the architecture, engineering, and construction (AEC) industry as it supports robotic understanding of the scene. This step is foundational for navigation as it provides a structured understanding of the environment's layout (Ekstrom and Hill 2023). For example, semantic recognition assigns meaningful labels to indoor spatial regions such as "kitchen" or "bedroom", which enriches the robot's situational awareness and facilitates context-specific interactions. Together, these capabilities form the backbone of smart assistive robots' ability to deliver personalized and adaptive mobility solutions.

In the construction phase of the built environment, scene understanding has been studied previously for site monitoring and progress tracking (Rao et al. 2022; Rebolj et al. 2017) material and equipment localization (Wang et al. 2019b), worker safety (Seo et al. 2015) and hazard detection (Jeelani et al. 2021). Most of these studies use convolutional neural networks (CNN), a widely used deep learning network, to perform object and room detection (LeCun et al. 2015). Recently, there have also been some emerging open-vocabulary point cloud detection methods, such as CLIP2Point (Huang et al. 2023), PointCLIP (Zhang et al. 2022). However, three major limitations in existing approaches highlight the research gaps addressed in this study and motivate the need for innovation:

First, CNN-based scene understanding methods operate within a closed vocabulary setting, which limits their adaptability to new and unseen categories of objects and environments (He et al. 2022). This limitation significantly impacts the functionality of assistive robots for people with disabilities, as these robots must interpret user instructions flexibly and adapt to varying environments (Tang et al. 2022). For instance, a closed-vocabulary system might successfully identify predefined places like "kitchen" or "bedroom" but fail to recognize semantically equivalent or descriptive instructions such as "a place to cook" or "a place to sleep", respectively. This rigidity restricts the robot's ability to assist in scenarios where environments are labeled or understood differently by users, such as multi-functional spaces or areas not explicitly defined by standard categories. Addressing this constraint is essential for developing assistive robots that can better understand and respond to diverse user needs, ensuring effective and intuitive support in real-world settings.

To mitigate this limitation, our framework introduces an open-vocabulary place segmentation and recognition pipeline that integrates vision-language models and large language models, enabling greater adaptability and semantic understanding. By "*open-vocabulary*", we mean that the models can classify objects beyond pre-defined categories.

Second, the current open-vocabulary point cloud understanding methods are not equipped to handle large-scale scene understanding, where a single point cloud can range from 1 million to tens of millions of points. Most existing approaches, such as CLIP2Point (Huang et al. 2023) and PointCLIP (Zhang et al. 2022), due to the lack of large-scale scene 3D-text datasets, rely on transformer-based models and relatively small 3D–text datasets to train feature extractors for object-scale point clouds, typically containing only 5,000 to 10,000 points. This scale mismatch presents two significant practical challenges: 1) Memory usage and computational overhead grow dramatically when attempting to process millions of points at once; the transformer structures used in these methods often have quadratic complexity, making large-scale inference prohibitively expensive (Zhou et al. 2024). 2) Large-scale point clouds often contain rich contextual information (such as spatial relationships, occlusions, and various object categories in a single environment) that cannot be captured by models trained solely on small-scale data (Liu et al. 2024).

To address the challenges of open-vocabulary understanding in large-scale point clouds, our method avoids the use of 3D transformer-based approaches. Instead, we adopt a two-stage strategy: 2D floor plan segmentation for precise room delineation and a 2D snap-lookup module for scene understanding, leveraging internet-scale pre-trained Vision Language Models (VLMs) to identify different rooms. By effectively transforming the 3D scene understanding problem into a 2D image understanding task, we mitigate the limitations of 3D point cloud processing.

Another key challenge in assistive robotics is managing the inherent uncertainty in open-vocabulary scene recognition and prediction tasks. Most existing methods rely on selecting the single result with the highest probability, often without considering the associated uncertainty (Czerniawski and Leite 2020; Lu et al. 2022; Wang et al. 2021). This approach, while seemingly straightforward, is prone to ambiguity and low success rates, particularly in complex or ambiguous environments. For example, when tasked with identifying a location such as "a place to rest" in a home with multiple bedrooms, conventional methods might select a single, highest-probability prediction bedroom that could be incorrect, leading to frustration or confusion for users.

To address this issue, our work integrates an improved version of conformal prediction as part of the uncertainty alignment process. Unlike traditional methods, Adaptive Conformal Prediction (ACP) provides statistical guarantees of prediction reliability by generating a prediction set rather than a single outcome. This ensures a more informed and flexible selection process, reducing ambiguity and improving overall decision-making in uncertain scenarios.

**Statement of Contribution**

In this work, we present a novel "Segment-Detect-Select" framework for open-vocabulary scene segmentation and classification, leveraging the open-vocabulary capabilities of VLM and the uncertainty quantification provided by our proposed ACP mechanism. The proposed framework begins with a 3D indoor scene point cloud, which is processed through our segmentation pipeline to extract distinct regions. Subsequently, images of these regions are captured, and the VLM computes similarity scores between user-provided natural language instructions and the visual representations of the regions. Finally, the framework identifies and selects regions that meet or exceed a specified threshold, as determined by the ACP, ensuring reliable and adaptive decision-making in open-vocabulary place recognition. These contributions enable assistive robots to interpret natural language instructions and adaptively recognize and localize semantically meaningful areas, such as "somewhere to sleep" or "a reading area" in homes and buildings, enhancing their ability to support users in complex, personalized indoor environments.

The specific contributions of our work are described below:

- We propose a novel "Segment-Detect-Select" open-vocabulary scene segmentation and classification framework designed specifically for assistive mobile robot scene understanding.
- We introduce a VLM-driven open-vocabulary segmentation and detection pipeline that achieves SoTA performance in comprehending highly complex, real-world environments.
- We present an ACP-based place selection framework for VLM place recognition, providing a statistical guarantee on uncertainty. This ensures a more informed and flexible decision-making process for selecting candidate places—such as bedrooms, corridors, or bathrooms— under varying environmental conditions and different people's instructions.
- We evaluated our framework on a widely used real-world Matterport3D dataset (MP3D), demonstrating that our approach outperforms existing methods in terms of segmentation (mIoU, AP50), detection (precision, recall et al.) and candidate selection (RmIoU).

## RELATED WORK

### 3D Semantic Scene Understanding

Some of the earliest efforts in semantic scene understanding for construction environments relied heavily on geometry-based methods to identify and label objects and places (Lee et al. 2013; Son et al. 2014). These methods typically fit simple geometric primitives (e.g., cylinders, planes) to point cloud data, enabling segmentation and classification based on shape or structural features. Techniques like RANSAC and machine learning classifiers have been employed to improve the robustness of object recognition and reduce noise in raw scans (Kyriakaki-Grammatikaki et al. 2022; Xu et al. 2017; Jung et al. 2016). Additionally, some studies utilize semantic Building Information Modeling (BIM) data to match real-world observations with corresponding elements in computer-aided design (CAD) models, aiding in more accurate scene segmentation and object detection (Chen et al. 2014; Cho and Gai 2014). Despite these advancements, geometry-based approaches often struggle with complex, irregularly shaped objects and environments that do not conform to simple primitives. Furthermore, attempts to match simulated or CAD-based objects with their real-world counterparts can fail due to variations in sensor data, material properties, or environmental conditions, thereby reducing reliability (Chen et al. 2019).

Recent work has focused on leveraging deep learning methods to directly map 3D inputs to object and place labels. In computer vision, deep learning has consistently demonstrated SoTA performance in scene classification and recognition tasks (LeCun et al. 2015). With sufficient data and training, deep learning models can robustly extract features from complex point clouds—encompassing diverse shapes and colors—thus enhancing scene understanding (Wang and Huang 2022). Some of the work utilizes 3D CNN-based structure to extract 3D point cloud features. Similar to 2D image classification, point cloud classification first generates a global embedding with an aggregation encoder and then passes the feature through several fully connected layers to obtain the final result (Lu and Shi 2020). There are some examples, such as 3D ShapeNet (Wu et al. 2015), and VoxelNet (Zhou and Tuzel 2018), that discretize the 3D space into a fixed-dimensional voxel grid and use 3D CNN to extract features and get labels. Moreover, some studies directly use point clouds as input to do the classification task. For example, PointNet (Qi et al. 2018)proposes the use of CNN to derive features from each point and a single max-pooling layer to aggregate features from all points. PointNet++ (Qi et al. 2017) builds based on PointNet and learns hierarchical features from several groups of points. Although these methods generally perform well in close vocabulary scene understanding, due to the limited labels they can differentiate, it is hard to adapt to different people's instructions in real-world settings.

Our framework leverages VLMs to segment and detect places in open-vocabulary settings, enabling the system to adapt seamlessly to novel environments and understand different user instructions.

**Open Vocabulary Scene Understanding**

Open-vocabulary detection and segmentation enable a model to learn from visual patterns containing novel objects that lack explicit annotations (Zhu and Chen 2024). Recently, there is significant progress that has been made in 2D image open-vocabulary understanding such as CLIP (Radford et al. 2021), BLIP-2 (Li et al. 2023), Grounding DINO (Liu et al. 2023) and Segment Anything (SAM) (Kirillov et al. 2023), largely due to the availability of internet-scale 2D-text datasets (Zhu and Chen 2024). CLIP (Radford et al. 2021) bridges vision and language by learning joint embeddings from internet-scale image-text pairs, enabling open-vocabulary recognition and zero-shot transfer across diverse visual tasks. BLIP-2 (Li et al. 2023) introduces a lightweight and efficient vision-language pretraining framework that bridges vision encoders and large language models. Grounding DINO (Liu et al. 2023) introduces language to closed-set detectors, enabling open-set object detection with human inputs such as category names or referring expressions. SAM (Kirillov et al. 2023) introduces a foundation model for image segmentation, enabling promptable segmentation tasks across diverse datasets. However, while 2D open-vocabulary models achieve remarkable performance, 3D point cloud open-vocabulary scene understanding remains underdeveloped due to the scarcity of large-scale 3D-text datasets (Huang et al. 2024).

In the 3D domain, existing point cloud understanding methods include CLIP2Point (Huang et al. 2023), PointCLIP (Zhang et al. 2022), OV-3DET (Lu et al. 2023) and OpenScene (Peng et al. 2023), which use pre-trained 3D VLM to detect and understand different scenes. CLIP2Point transfers CLIP's vision-language pre-training to point cloud classification by introducing an image-depth pre-training method. PointCLIP (Zhang et al. 2022) extends CLIP's zero-shot and few-shot capabilities to 3D recognition by projecting point clouds onto multi-view depth maps and aligning them with 3D category texts. OV-3DET (Lu et al. 2023) proposes a dividing-and-conquering strategy for open-vocabulary 3D detection without 3D annotations, leveraging image pre-trained models and cross-modal contrastive learning. OpenScene (Peng et al. 2023) predicts dense CLIP features for 3D scene points, enabling zero-shot semantic segmentation and open-vocabulary queries for various scene understanding tasks. Most of these methods utilize 3D transformer-based methods for object detection and segmentation. However, unlike 2D - text datasets such as COCO-text (Veit et al. 2016) and ICDAR (Karatzas et al. 2013), where we have internet-scale data, 3D-text datasets such as ScanRefer (Chen et al. 2020) and SQA3D (Ma et al. 2023) are very limited and most of them are limited to 3D object level instead of scene level and it is hard for these models to generalize from object to scene-scale data.

In order to compensate for the limitation of 3D scene understanding, we propose a two-stage room segmentation and detection framework in which we segment the floor plan and utilize 2D VLM to detect different rooms. Our proposed framework addresses the limitations of 3D scene understanding by bridging the gap between 3D and 2D open-vocabulary tasks. By transforming the 3D point cloud problem into a 2D open-vocabulary understanding task, we leverage the strengths of internet-scale datasets and pre-trained VLM, which are more mature and versatile in the 2D domain.

**Hallucinations and Uncertainty in VLM**

VLMs face a critical challenge in their tendency to hallucinate. Hallucination refers to the generation of outputs that are not grounded in the input data, resulting in fabricated or overly confident information (Farquhar et al. 2024). This issue can result in misidentifying locations with high confidence, leading to navigation errors and compromised contextual understanding, particularly in assistive applications (Jha et al. 2023; Zhang et al. 2023). Additionally, natural language instructions provided in real-world scenarios often contain inherent or unintentional ambiguity, further exacerbating the likelihood of hallucinations (Hatori et al. 2018). For instance, when a robot is instructed to say "I am sleepy" in a home environment, the ambiguity in the instruction can cause errors in selecting the wrong room. Rather than acting in such uncertain situations and risking incorrect behavior, the robot should recognize its uncertainty and select all of the rooms that fit the instructions.

Recent research has focused on quantifying and mitigating hallucinations and uncertainty in VLMs and Large Language Models (LLMs). Semantically Diverse Language Generation (SDLG) (Aichberger et al. 2024) has been shown to measure predictive uncertainty and detect potential hallucinations. (Yadkori et al. 2024) developed an information-theoretic metric to detect high epistemic uncertainty, which can indicate unreliable model outputs and hallucinations. (Du et al. 2023) proposed an association analysis approach to quantify hallucination levels and investigate their underlying causes, revealing potential deficiencies in LLMs' capabilities. Some recent studies attempt to handle hallucinations by using conformal prediction-based methods to provide coverage guarantee (Kumar et al. 2023). Know When Robot Don't Know (KnowNo) (Ren et al. 2023a) uses conformal prediction (CP) to create coverage guarantees for robotics tasks, asking users for help when the agent is unsure. Seeing with Partial Certainty (SwPC) (Xu et al. 2025) adapts CP to open-vocabulary place segmentation and detection tasks, allowing users to provide help when hallucinations and uncertainty occur.

However, as discussed in (Angelopoulos and Bates 2023), conformal prediction struggles to adapt to varying data sizes because it relies on raw cosine similarity scores as the non-conformity measure. When data sizes differ, the scale of the cosine similarity scores for rooms matching the user's description also changes. For instance, in an environment with two rooms that meet the description, the similarity score thresholds are typically on the order of $10^{-1}$ (e.g., each room has a similarity score of 0.4, other rooms are below 0.2). However, in an environment with 10 rooms and about matching rooms, the similarity scores decrease to the order of $10^{-2}$ (0.095 for each room and the sum of other rooms is 0.05). This imbalance can impact the model's reliability in diverse real-world environments, particularly in high-stakes applications like assistive robotics, where robust uncertainty handling across all subgroups is critical.

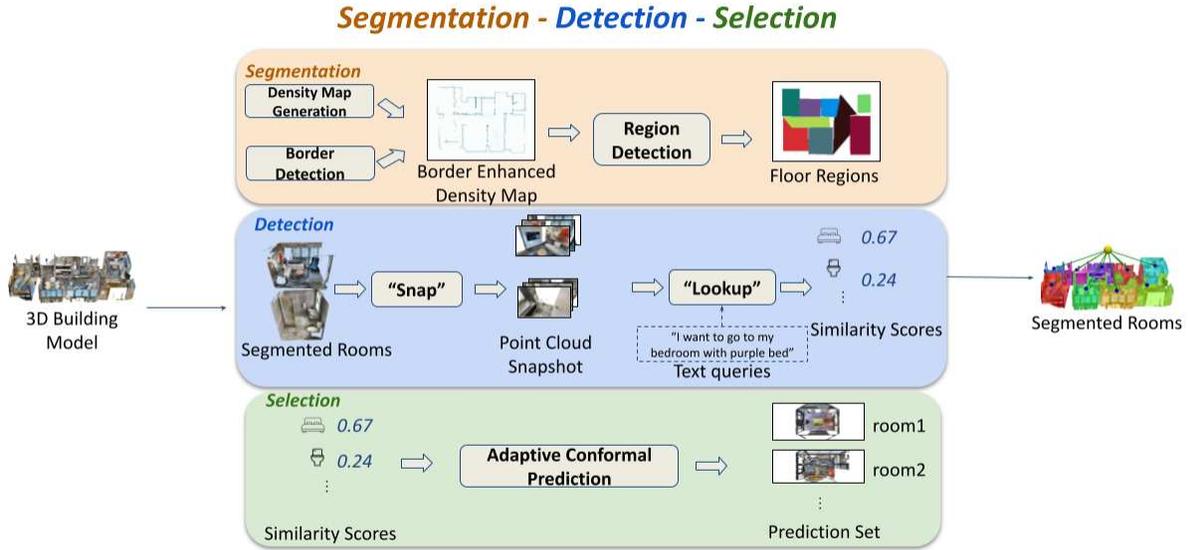

**Fig. 1.** The general system pipeline

To mitigate this issue, we introduce ACP into the uncertainty alignment framework for VLM place recognition. Different from CP, ACP can dynamically change prediction set sizes according to the environment room sizes so that we can provide a more accurate selection result than CP.

## RESEARCH METHODOLOGY

The entire VLM-based place segmentation and recognition pipeline are illustrated in Fig. 1. This pipeline divides the scene understanding task into three distinct stages: **Segmentation, Detection, and Selection**. (1) The **Segmentation** module combines geometric methods to enhance the density map's boundaries and employs a learning-based approach to detect and isolate floor regions; (2) The **Detection** module comprises two submodules: the "snap" module captures multiple images of the room, while the "lookup" module leverages a pre-trained 2D VLM to compute similarity scores between user-provided text queries and the captured snapshots; and (3) Based on the computed similarity scores, the **Selection** module uses ACP to identify and select rooms that satisfy the criteria.

**Segmentation Module**

In order to segment each room accurately, as shown in Fig. 1, we propose BorderFormer, where we apply a geometry-based method to generate a border-enhanced density map. The border detection method is shown in Fig.2.

First, the input point cloud is segmented into $N$ slices along the z-axis, with each slice projected onto an occupancy grid map denoted as $G_k, k = 1, ..., N$. We calculate the free space area of each grid map, denoted as $S_k, k = 1, ..., N$. To reduce noise, we eliminate grid maps that lack sufficient border information or contain excessive noise. The detailed coverage criteria are outlined in (1).

$$\mathbf{G}_{\text{select}} = \{G_k \mid \delta_b S < S_k < \delta_t S, \ k = 1, ..., N\} \tag{1}$$

where $\mathbf{G}_{\text{select}}$ is the selected set of projected grid maps. The parameters $\delta_b$ and $\delta_t$ are empirically chosen as $\frac{1}{15}$ and $\frac{1}{5}$, respectively. These selected grid map layers are then merged to construct the border map denoted as $G_{border}$, shown in (2):

$$G_{border}(i, j) = \begin{cases} 1 & \text{if } \sum_{k=1}^{M} G_k(i, j) \geq \frac{3}{4}M \\ 0 & \text{otherwise} \end{cases}. \tag{2}$$

where $M$ represents the number of valid grid maps after elimination. This step ensures that only the regions consistently identified across most layers are considered wall boundaries.

To generate the border-enhanced density map $G_{combine}$, the border map is combined with the density map $G_{den}$ projected directly from the original point cloud:

$$G_{combine} = \gamma G_{den} + (1 - \gamma) G_{border} \tag{3}$$

where $\gamma$ is a weight parameter that is empirically chosen as 0.9. This process emphasizes the wall boundaries, facilitating accurate region detection.

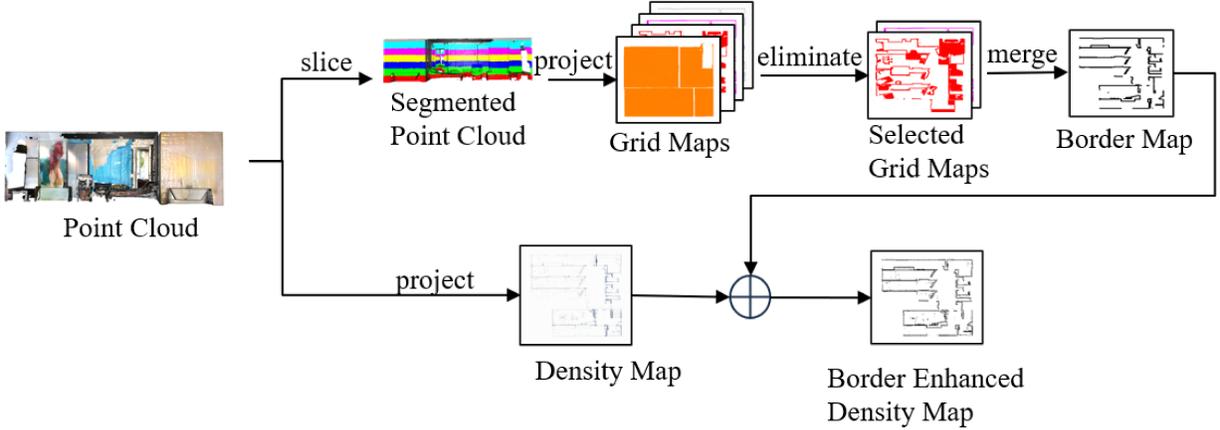

**Fig. 2.** Generation of border-enhanced density map

For region detection, we employed the latest SoTA method, RoomFormer (Yue et al. 2023), as the region detector. Building on the original training provided by the paper on 3,000 scenes from Structure3D (Zheng et al. 2020), we further fine-tuned the model with 100 scenes from Matterport3D (MP3D) dataset (Chang et al. 2017).

**Detection Module**

After we obtain the segmented rooms, inspired by (Huang et al. 2024), we apply a "Snap-Lookup" pipeline to get the room label for each room in open-vocabulary settings. To get the 2D image for VLM to identify the types, we apply a "Snap" module as shown in Fig. 3a. We designed the camera pose $p_{camera} = (x, y, z)$ shown in (4):

$$\frac{4(x - x_c)}{L^2} + \frac{4(y - y_c)}{W^2} = 1$$
$$z = z_c$$
(4)

where the room's length and width is $(L, W)$ and its center at $p_{center} = (x_c, y_c)$. We evenly position the camera in an elliptical trajectory around the room and let the camera face the center to take snapshots from different positions.

To obtain open-vocabulary features for each room, we adopt the approach from (Werby et al.

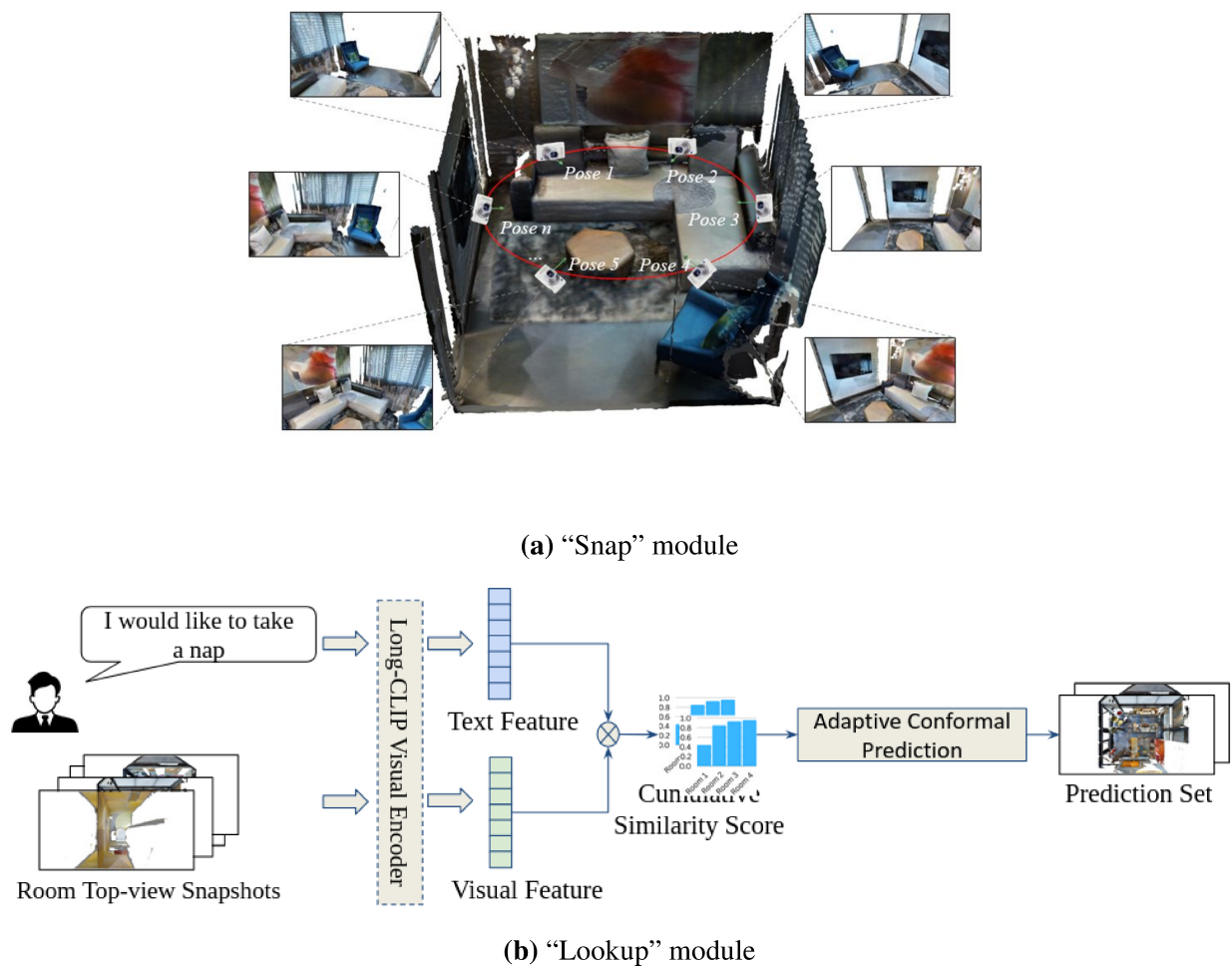

**(a)** "Snap" module

**(b)** "Lookup" module

**Fig. 3.** "Snap" and "Lookup" modules in room open-vocabulary classification

2024) and leverage the Long-CLIP (Zhang et al. 2025) visual encoder to extract embeddings from both images and user instructions. Since some images may be captured from suboptimal angles, potentially leading to the misclassification of room types, we refine these features by selecting K-representative view embeddings using the K-means algorithm. We then compute a cosine similarity matrix between the K-representative features and the text features extracted by Long-CLIP (Zhang et al. 2025). This process is illustrated in Fig. 3b within the "Lookup" module. Finally, the computed similarity scores are input into the "Selection" module, which identifies the rooms that best match the user's requirements.

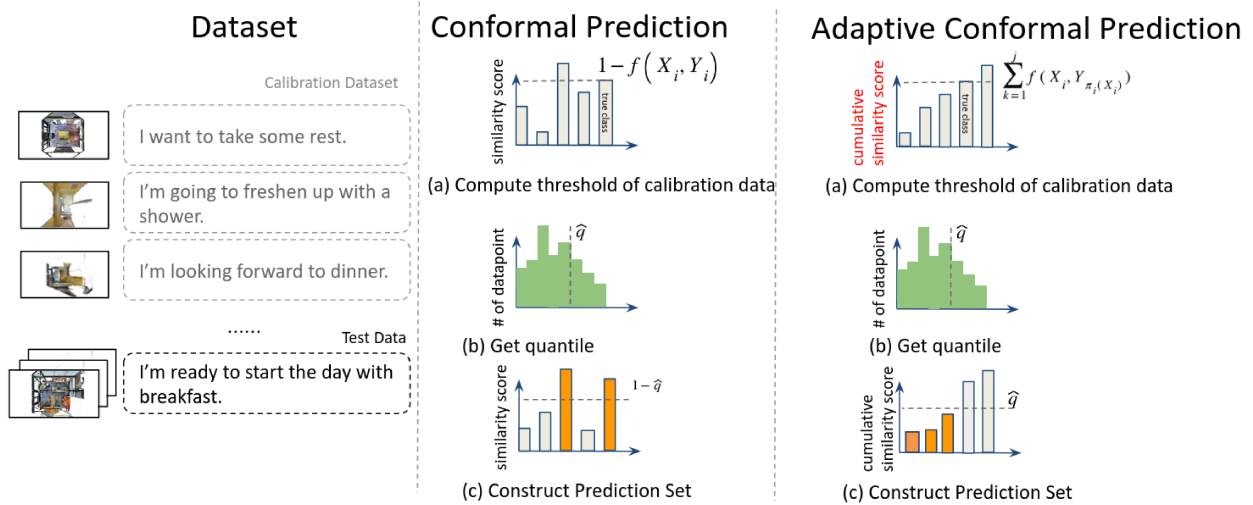

**Fig. 4.** Comparison between conformal Prediction and Adaptive Conformal Prediction

**Selection Module**

After computing the similarity scores, the selection process begins. Due to the inherent ambiguity in natural language instructions and potential hallucinations in VLM, multiple rooms may match a user's needs. Instead of relying on traditional methods that simply select the room with the highest similarity score, we propose a novel approach: Adaptive Conformal Prediction (ACP). Compared with the conformal prediction (CP) proposed earlier (Xu et al. 2025), our proposed ACP can adapt the prediction to different datasets with different sizes by using the cumulative similarity score and can set up different criteria to select candidates for the prediction dataset. The difference between CP and ACP is shown in Fig. 4. The proposed ACP algorithm is separated into two steps: data calibration and prediction set generation. The ACP steps are illustrated below.

*Calibration Dataset Collection*

We first need to obtain the calibration dataset to calibrate the uncertainty. The steps of construction of the calibration dataset are below:

1. We utilize the fine-tuned VLM to get the cosine similarity score between different descriptions and room images.

2. We sort the cosine similarity score and form the non-conformity score - image - description

pairs as the raw datasets used for ACP calibration. Inspired by the oracle algorithm (Gibbs and Candès 2024), the non-conformity score is described as the cumulative similarity score. We define $\pi(X_i) = \{\pi_1(X_i), ..., \pi_n(X_i)\}$ to be the permutation of $\{1, ..., n\}$ that sorts $\{f(X_i, Y_1), ..., f(X_i, Y_n)\}$ from most likely to least likely. The dataset is described below:

$$\mathcal{D} = \{(s(X_i, Y_j), X_i, Y_j) | s(X_i, Y_j) = \sum_{k=1}^{j} f(X_i, Y_{\pi_i(X_i)}), i = 1..., n, j = \pi_1(X_i), ..., \pi_n(X_i)\} \tag{5}$$

3. We incrementally include classes in our set until the true label is reached, stopping at that point. Unlike the standard CP described in (Xu et al. 2025), this approach leverages the softmax output of the similarity scores between people's instruction and all different rooms. According to the theory of conformal prediction, we form the calibration dataset as below:

$$\hat{\mathcal{D}} = \{(s(X_i, Y_j), X_i, Y_j) | (s(X_i, Y_j), X_i, Y_j) \in \mathcal{D}, i = 1, ..., n, j = \pi_1(X_i), ..., \pi_i(X_i)\} \tag{6}$$

*Optimal Error Rate Calculation*

After the calibration dataset is obtained, we can obtain the optimal error rate $\alpha^*$ by solving an optimization problem over the validation set of the MP3D dataset. We will perform adaptive conformal prediction over different $\alpha$ to get our prediction set. Similar as the step 4 of calibration dataset collection, we define $\pi(X_{val}) = \{\pi_1(X_{val}), ..., \pi_n(X_{val})\}$ to be the permutation of $\{1, ..., n\}$ that sorts $\{f(X_{val}, Y_1), ..., f(X_{val}, Y_n)\}$ from most likely to least likely. Then, we define a score function as below:

$$s(X_{val}, Y_{\pi_k(X_{val})}) = \sum_{j=1}^{k} f(X_{val}, Y_{\pi_j(X_{val})}), \text{where } k = 1, ..., n \tag{7}$$

Next step is to set the quartile value $\hat{q}$ of the calibration dataset which is the same as in any conformal prediction.

$$\hat{q} = \text{Quartile}(\hat{\mathcal{D}}, \frac{\lceil (n+1)(1-\alpha) \rceil}{n}) \tag{8}$$

where $\hat{D}$ is the calibration dataset that is from the previous calibration dataset generation step. $\alpha \in [0, 1]$ is the error rate. To obtain the optimal error rate, we will form the optimization problem as below:

$$\alpha^* = \arg \max_{\alpha \in [0,1]} \frac{|C(X_{val}) \cap C_{true}(X_{val})|}{|C(X_{val}) \cup C_{true}(X_{val})|}$$
$$s.t. C(X_{val}) = \{Y_{\pi_1(X_{val})}, ..., Y_{\pi_k(X_{val})}\}, \qquad (9)$$
$$\text{where } k = \sup\{k' : s(X_{val}, Y_{\pi_{k'}(X_{val})}) \leq \hat{q}\} + 1$$

where $C(X_{val})$ is calculated using ACP and $C_{true}(X_{val})$ is the ground truth prediction set that contains all rooms that fit the room's description.

*Adaptive Conformal Prediction*

For people who provide the room description $X_{people}$, we will perform adaptive conformal prediction using the optimal error rate $\alpha^*$ obtained from the previous steps to get our prediction set. Simliar to the previous section, we define $\pi(X_{people}) = \{\pi_1(X_{people}), ..., \pi_n(X_{people})\}$ to be the permutation of $\{1, ..., n\}$ that sorts $\{f(X_{people}, Y_1), ..., f(X_{people}, Y_n)\}$ from most likely to least likely. Then, we define a score function as below:

$$s(X_{people}, Y_{\pi_k(X_{people})}) = \sum_{j=1}^{k} f(X_{people}, Y_{\pi_j(X_{people})}), \text{ where } k = 1, ..., n \qquad (10)$$

The next step is to set the quartile value $\hat{q}$ of the calibration dataset, which is the same as in any conformal prediction.

$$\hat{q} = \text{Quartile}(\hat{D}, \frac{\lceil (n+1)(1-\alpha^*) \rceil}{n}) \qquad (11)$$

where $\hat{D}$ is the calibration dataset that is from the previous calibration dataset generation step. $\alpha^* \in [0, 1]$ is the optimal error rate. Then we perform ACP to obtain the final prediction dataset.

$$C(X_{people}) = \{Y_{\pi_1(X_{people})}, ..., Y_{\pi_k(X_{people})}\},$$
$$\text{where } k = \sup\{k' : s(X_{people}, Y_{\pi_{k'}(X_{people})}) \leq \hat{q}\} + 1 \qquad (12)$$

# EXPERIMENTAL RESULTS

In this section, we evaluate our proposed **Segmentation - Detection - Selection** framework in the following three parts respectively. The evaluation results are presented below:

**Dataset**

We evaluate our **Segmentation - Detection - Selection** module in the Matterport3D (MP3D) dataset (Chang et al. 2017). The MP3D dataset is a comprehensive large-scale collection of indoor scene data captured using advanced 3D scanning technology, providing a rich multi-modal resource that includes high-resolution RGB images, corresponding depth maps, and detailed 3D mesh reconstructions taken from multiple viewpoints to offer full panoramic context. Covering over 90 diverse indoor environments ranging from residential homes and apartments to offices and commercial spaces, the dataset captures the complexity and variability of real-world settings, making it an ideal benchmark for tasks in computer vision, robotics, augmented reality, and 3D reconstruction. Each scene is annotated with detailed semantic labels, enabling precise tasks such as semantic segmentation, object recognition, and spatial understanding, while the inclusion of panoramic imagery supports comprehensive scene analysis and indoor navigation research.

**Segmentation Module Evaluation**

We conducted our experiment on the MP3D dataset, where all multi-floor scenes were segmented into single-floor scenes, resulting in a total of 189 scenes. Excluding those used for fine-tuning the region detector, we selected 43 scenes to serve as the test set for room segmentation. We compared our method to RoomFormer (Yue et al. 2023), the current SOTA in learning-based algorithms, and the room segmentation techniques employed in HOV-SG (Werby et al. 2024), the SOTA in geometry-based algorithms. Our evaluation metrics included Average Precision at 50% overlap (AP50) and mean Intersection over Union (mIoU) (Werby et al. 2024).

In our experimental results, shown in TABLE 1, by generating a border-enhanced density map before input to RoomFormer, our proposed BorderFormer approach achieved 12% improvements in AP50 and 3% in mIoU. This demonstrates the potential of our method as an effective pre-processing module for room segmentation. The qualitative comparison of room segmentation is shown in Fig. 5, showcasing the superior performance of the room segmentation pipeline.

| Methods | AP50 | mIoU |
|---|---|---|
| HOV-SG (Werby et al. 2024) | 0.06 | 0.23 |
| RoomFormer (Yue et al. 2023) | 0.41 | 0.56 |
| **BorderFormer** (proposed method) | **0.53** | **0.59** |

**TABLE 1.** Comparison of different room segmentation methods on the MP3D dataset.

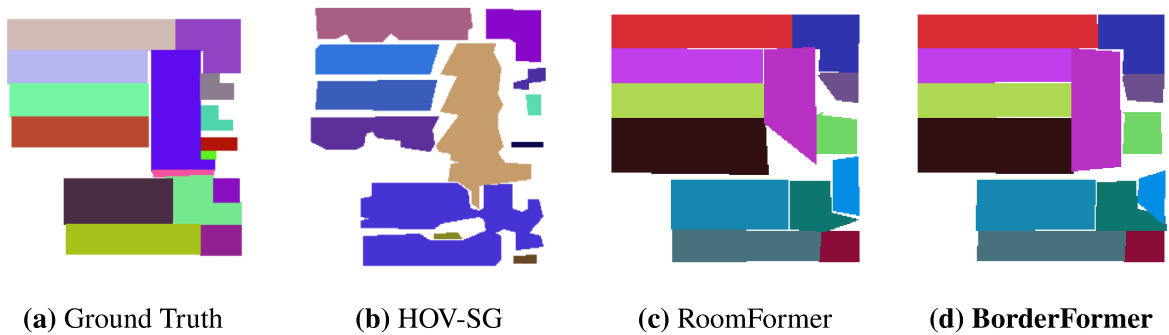

(a) Ground Truth     (b) HOV-SG     (c) RoomFormer     (d) **BorderFormer**

**Fig. 5.** Qualitative result of room segmentation

**Detection Module Evaluation**

We evaluate our Snap-Lookup pipeline for room classification on 100 segmented room scenes from the MP3D dataset (Chang et al. 2017), using the complete set of room categories present in these scenes. As baselines, we compare two approaches: the zero-shot LLM-based room type inference method (Mehan et al. 2024) and the room classification approach from HOV-SG (Werby et al. 2024), a leading open-vocabulary scene graph generation method. These baselines are categorized into two types: privileged, which utilizes ground truth object types for LLM inference, and unprivileged, which relies on object types detected by our proposed object detection algorithm. We employ GPT-3.5-turbo (OpenAI 2023) and GPT-4 (OpenAI 2024) as the LLM baselines.

Our evaluation metrics include Precision, Recall, weighted F1 score, and mean Average Precision (mAP) (Powers 2020). In TABLE 2, Precision quantifies the accuracy of positive predictions, Recall measures the model's ability to identify all relevant instances, and F1 score represents the harmonic mean of Precision and Recall. Meanwhile, mAP evaluates the classification performance across all categories.

|  | Methods | Precision | Recall | F1 | mAP |
|---|---|---|---|---|---|
| Privileged | GPT-3.5-turbo w\ GT object | 0.69 | 0.61 | 0.61 | 0.63 |
|  | GPT-4o w\ GT object | 0.68 | 0.63 | 0.61 | 0.63 |
| Unprivileged | GPT-3.5-turbo w\ object detected | 0.10 | 0.14 | 0.10 | 0.14 |
|  | GPT-4o w\ object detected | 0.18 | 0.19 | 0.15 | 0.19 |
|  | HOV-SG (Werby et al. 2024) | 0.51 | 0.20 | 0.22 | 0.19 |
|  | **Snap-Lookup** (proposed method) | **0.69** | **0.70** | **0.66** | **0.69** |

**TABLE 2.** Comparison of room classification method on MP3D dataset.

The Snap-Lookup pipeline improves room classification by integrating visual features into the type inference process, enabling it to distinguish between rooms that share similar objects—an area where text-only inference methods like GPT-3.5-turbo and GPT-4 fall short. Additionally, our Snap module captures comprehensive snapshots of entire rooms, ensuring higher-quality image inputs without being constrained to specific viewpoints, unlike HOV-SG, which relies on predefined camera positions. As demonstrated in TABLE 2, Snap-Lookup achieves notable improvements across both privileged and unprivileged baselines, highlighting its effectiveness in open-vocabulary room classification.

**Selection Module Evaluation**

We evaluate our proposed **Selection** framework using ACP in a diverse set of indoor environments. In the experiment below, first, we have demonstrated the result of our selection of error rate $\alpha$. Then, we conducted the experiment across different people's instructions, which demonstrated the effectiveness of reaching the maximal mIoU.

We conducted our experiment on the MP3D dataset, where all multi-floor scenes were segmented into single-floor scenes, resulting in 189 scenes. Excluding those used for calibration of ACP, we selected 14 scenes, which include 400 rooms, to serve as the test set for room segmentation. Since we are using the same VLM model to generate the similarity score for the calibration set and test set, the test data and calibration data are independently and identically distributed.

*Optimal Error Rate Calculation*

In order to calculate the optimal error rate $\alpha$, we choose the validation dataset of MP3D to calculate mIoU over different error rates $\alpha \in [0, 1]$. The way to calculate mIoU is shown in (9). Noted $X_{val}$ refers to the class labels of MP3D, and $Y_{\pi_k(X_{val})}$ refers to different rooms of the scene. The result over different $\alpha$ is shown in Fig. 6. From the plot, the optimal value for ACP and CP is 0.3. We are using this error rate for the following evaluation.

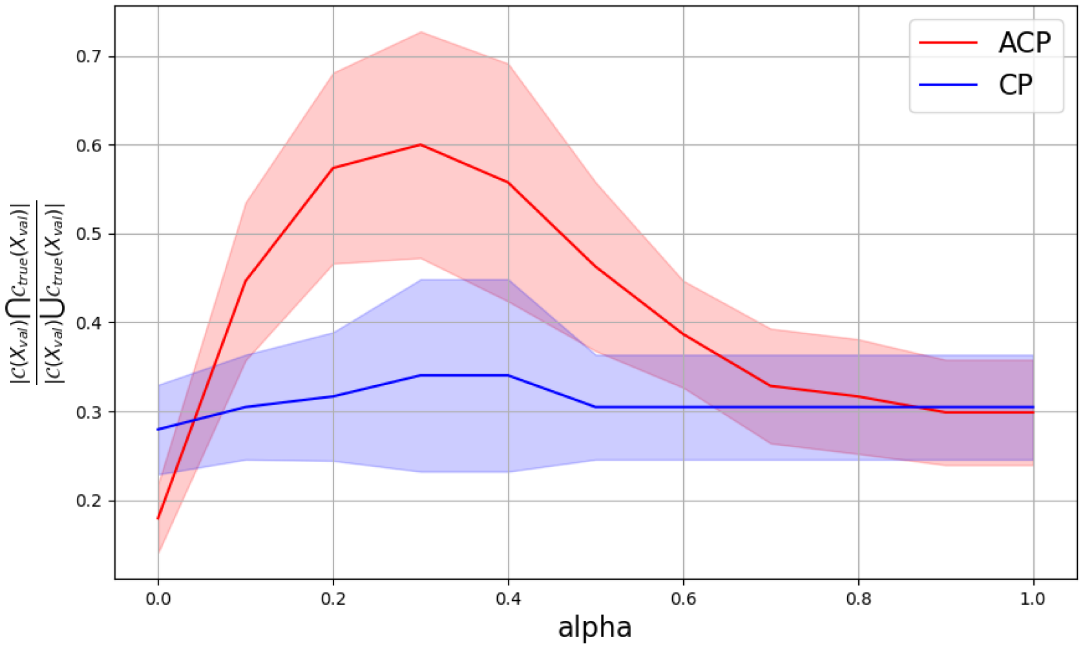

**Fig. 6.** Qualitative result of our proposed Selection Module

*Experimental Results*

Our baselines for comparison are the CP method as illustrated in the METHODOLOGY section, and Prompt Set. The introduction of these baselines is shown below:

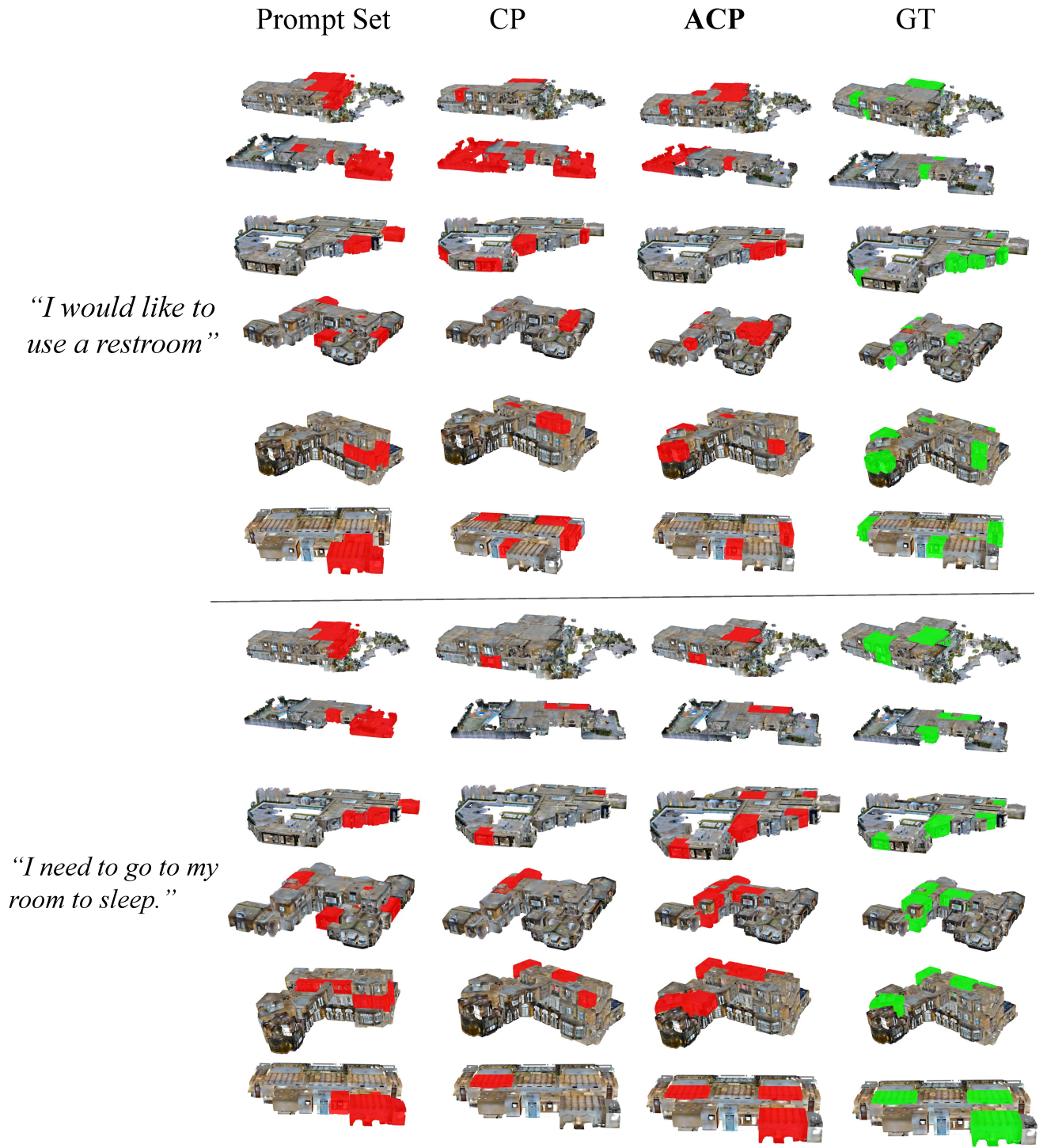

**Fig. 7.** Qualitative result of our proposed Selection Module

| Instructions | GT Room Type | Prompt | CP | **ACP** |
|---|---|---|---|---|
| *I'm ready to head to my bedroom and get some rest.* | bedroom | 0.1097 | 0.1274 | **0.3345** |
| *I need to go to my room to sleep.* | bedroom | 0.1447 | 0.1631 | **0.3208** |
| *I'm looking forward to retreating to my bedroom for a good night's sleep.* | bedroom | 0.0847 | 0.2310 | **0.3651** |
| *I want to escape to my room for some well-deserved rest.* | bedroom | 0.0922 | 0.1631 | **0.3241** |
| *I'm eager to hit the bedroom and catch some z's.* | bedroom | 0.0811 | 0.1631 | **0.2579** |
| *It's time to sleep.* | bedroom | 0.0725 | 0.1988 | **0.2731** |
| *I need to use the toilet and then take a shower.* | bathroom | 0.1061 | 0.1851 | **0.3000** |
| *I'm heading for a toilet.* | bathroom | 0.0883 | 0.1672 | **0.2827** |
| *I need to excuse myself to the restroom.* | bathroom | 0.1190 | 0.2786 | **0.3985** |
| *I'm going to freshen up with a shower.* | bathroom | 0.1303 | 0.2786 | **0.4428** |
| *I could use a bathroom break.* | bathroom | 0.1133 | 0.2565 | **0.3718** |
| *I feel like having something to eat.* | kitchen, dining room | 0.0850 | 0.1667 | **0.1788** |
| *I'm looking forward to dinner.* | dining room | 0.0643 | 0.1667 | **0.1709** |
| *I'm ready to start the day with breakfast.* | kitchen, dining room | 0.0607 | 0.1905 | **0.2223** |
| *I'm feeling hungry.* | kitchen, dining room | 0.0464 | 0.1667 | **0.1688** |
| *It's dinner time.* | dining room | 0.0464 | 0.1786 | **0.2363** |
| *Average* | | 0.0903 | 0.1926 | **0.2893** |

**TABLE 3.** RmIoU for different people's instruction over Prompt Set, CP and ACP

- **CP Set** (Xu et al. 2025): CP set utilizes conformal prediction to set the threshold value so that a statistical guarantee of a certain success rate is possible. It uses $1 - f(X_i, Y_i)$ as the non-conformity score to get the prediction set.

- **Prompt Set** (Ren et al. 2023b): Prompts the multimodal LLM to output the prediction set directly. (e.g. "Prediction set: [B, D]")

We evaluated our method with baselines over the MP3D testing dataset. To validate our assumption that our method can outperform other baselines in selecting rooms according to people's different expressions, we generate 15 different expressions using an LLM for different room types.

Since we want to let the selection process make a selection that contains as many correct room types and as few wrong room types as possible, drawing from the idea of mIoU (Taran et al. 2018), we evaluate the accuracy of our selection module using the Room mIoU (RmIoU) for each prompt

$X_{people}$ over the 14 scenes. The calculation of the selection module is shown below:

$$RmIoU = \frac{\sum_{i=1}^{N} \frac{|C(X_{people}) \cap C_{true}(X_{people})|}{|C(X_{people}) \cup C_{true}(X_{people})|}}{N} \quad (13)$$

where $C(X_{people})$ represents the selected rooms set using our method or baseline methods, and $C_{true}(X_{people})$ represents the ground truth room set corresponding to user's instructions.

As shown in Fig. 7, we compare the qualitative results of our proposed ACP-based selection "GT" means the ground truth room type. As we can see from the figure, we test the selection module on two different prompts with error rate $\alpha$ to be set at 0.3. From the qualitative result, our proposed ACP-based selection methods yield the best result compared to other methods. We evaluate the RmIoU for the two baseline methods and our method. The result is shown in TABLE 3. In the table, "Instructions" refers to the different instructions for one particular room type. "GT Room Type" means the ground truth room type corresponding to the instructions. From the result, we know that the average RmIoU of our methods outperforms the Prompt Set by 19.9% (p<0.001) and outperforms the CP set by 9.67% (p<0.001). Since ACP uses the cumulative cosine similarity score as the non-conformity score, it can adapt to different data sizes and do better uncertainty alignment than CP. Regarding the Prompt Set, the VLM's tendency to hallucinate and exhibit bias results in the lowest success rate.

## CONCLUSION

In this work, we presented an open-vocabulary semantic segmentation and scene recognition framework designed to enhance assistive robot navigation in complex indoor environments. Our Segment-Detect-Select pipeline integrates Vision-Language Models (VLMs) and Large Language Models (LLMs) to enable adaptive and intuitive scene understanding. By leveraging a multi-stage approach—segmenting spatial regions, detecting semantic labels, and selecting the most relevant spaces using Adaptive Conformal Prediction (ACP)—our framework overcomes the limitations of traditional closed-vocabulary methods and enhances robustness in uncertain environments. In the Segmentation module, our proposed BorderFormer approach achieves a 12% improvement in AP50 and a 3% improvement in mIoU. In the Detection module, our Snap-Lookup method outperforms the current state-of-the-art by 18% in Precision, 50% in Recall, 44% in F1 Score, and 50% in mAP under unprivileged conditions. Finally, for the Selection module, our approach achieves an average RmIoU that surpasses the Prompt Set by 19.9% and the CP set by 9.67%.

Practically, this capability allows assistive robots—such as smart wheelchairs—to interpret a wide range of natural language commands from users, including vague or personalized expressions like "*I'm feeling hungry*" or "*It's time to sleep.*" This leads to more meaningful and flexible interactions, reduces the cognitive and physical burden on users, and improves trust and satisfaction with assistive technologies. It also makes the robot more responsive to the unique and dynamic ways people refer to spaces in their homes or workplaces, which is critical for enabling autonomy for people with disabilities in real-world settings.

Beyond assistive navigation, the proposed framework has broad applicability in domains such as search-and-rescue operations in unknown buildings and exploratory robotics in unfamiliar indoor settings, where interpreting high-level spatial semantics from human instructions is essential. By bridging the gap between large-scale 3D environments and 2D open-vocabulary vision-language understanding, this work lays the groundwork for a new generation of socially aware, semantically intelligent, and uncertainty-aware robotic systems.

## DATA AVAILABILITY STATEMENT

Some data, models, or code that support the findings of this study are available from the corresponding author upon reasonable request.

## ACKNOWLEDGEMENTS

The work presented in this paper was supported financially by the United States National Science Foundation (NSF) via Award# SCC-IRG 2124857. The support of the NSF is gratefully acknowledged.